\pdfoutput=1
\documentclass[11pt]{article}

\usepackage{acl}

\usepackage{times}
\usepackage{latexsym}
\usepackage{amsmath}
\usepackage{subcaption}
\usepackage{newtxtext,newtxmath}
\usepackage{graphicx}
\usepackage{booktabs}
\usepackage{amsfonts}

\usepackage[T1]{fontenc}
\usepackage[utf8]{inputenc}

\usepackage{microtype}

\title{Linear Transformers with Learnable Kernel Functions are Better In-Context Models}

\author{
Yaroslav Aksenov\textsuperscript{\rm $\clubsuit$ \rm $\vardiamondsuit$}, 
Nikita Balagansky\textsuperscript{\rm $\clubsuit$ \rm $\varheartsuit$}, 
Sofia Maria Lo Cicero Vaina\textsuperscript{\rm $\spadesuit$}\thanks{Work done while at Tinkoff.}, \\ 
\textbf{Boris Shaposhnikov\textsuperscript{\rm $\clubsuit$}, 
Alexey Gorbatovski\textsuperscript{\rm $\clubsuit$}, 
Daniil Gavrilov\textsuperscript{\rm $\clubsuit$}} \\
    \textsuperscript{\rm $\clubsuit$}Tinkoff 
    \textsuperscript{\rm $\vardiamondsuit$}Higher School of Economics\\
    \textsuperscript{\rm $\varheartsuit$}Moscow Institute of Physics and Technology 
    \textsuperscript{\rm $\spadesuit$}Innopolis University\\ 
    \textit{n.n.balaganskiy@tinkoff.ru}
}

\newcommand{\specialcell}[2][c]{%
  \begin{tabular}[#1]{@{}r@{}}#2\end{tabular}}

\begin{document}
\maketitle
\begin{abstract}
Advancing the frontier of subquadratic architectures for Language Models (LMs) is crucial in the rapidly evolving field of natural language processing. Current innovations, including State Space Models, were initially celebrated for surpassing Transformer performance on language modeling tasks. However, these models have revealed deficiencies in essential In-Context Learning capabilities -- a domain where the Transformer traditionally shines. The Based model emerged as a hybrid solution, blending a Linear Transformer with a kernel inspired by the Taylor expansion of exponential functions, augmented by convolutional networks. Mirroring the Transformer's in-context adeptness, it became a strong contender in the field. In our work, we present a singular, elegant alteration to the Based kernel that amplifies its In-Context Learning abilities evaluated with the Multi-Query Associative Recall task and overall language modeling process, as demonstrated on the Pile dataset. 
\end{abstract}

\section{Introduction}

Large Language Models (LLMs) are revolutionizing the field of natural language processing and establishing new benchmarks across various tasks \citep{llama2, mistral}. Nevertheless, despite their triumphs, most of these models are built on Transformer frameworks that employ attention mechanisms. These mechanisms scale poorly with long text sequences, leading to impractical computational complexity for extending contextual processing \citep{transformer, lra}.

To address this constraint, several alternatives to Transformers were proposed. \citet{linear} suggested replacing the exponential function in the attention mechanism with the kernel function to change the order of computations and thus move away from quadratic complexity of the sequence length. However, when compared to vanilla Transformers, this approach leads to a drop in performance. Furthermore, the kernel function selection is a topic still in need of consideration. An alternative way to define a linear model is to utilize State Space Models (SSMs) \citep{s4, s5, mamba}, which are capable of producing quality that is comparable to Transformers when measured with perplexity on language modeling. 

Notably, both Linear Transformers \citet{linear} and SSMs can be described as Recurrent Neural Networks (RNNs) \citep{gru, lstm}, which have their limitations when it comes to managing lengthy dependencies within texts since memory capacity can be overrun as the volume of information increases. Additionally, while the hidden state of RNNs is larger for Linear Transformers than for SSMs, the latter showed higher text modeling quality. The introduction of the Based model \citep{based} attempted to address the abovementioned challenges by utilizing a hybrid architecture \citep{h3} based on a Linear Transformer with a novel kernel function derived from a Taylor expansion of an exponential function. \citet{based} demonstrated that the Based model was less prone to performance issues when working with longer content than other models when assessed on the Multi-Query Associative Recall (MQAR) task. Nonetheless, even the Based model experiences a drop in performance when faced with extensive contexts relative to the conventional transformer architecture.

\begin{figure*}[t]
    \centering
    \includegraphics[width=\linewidth]{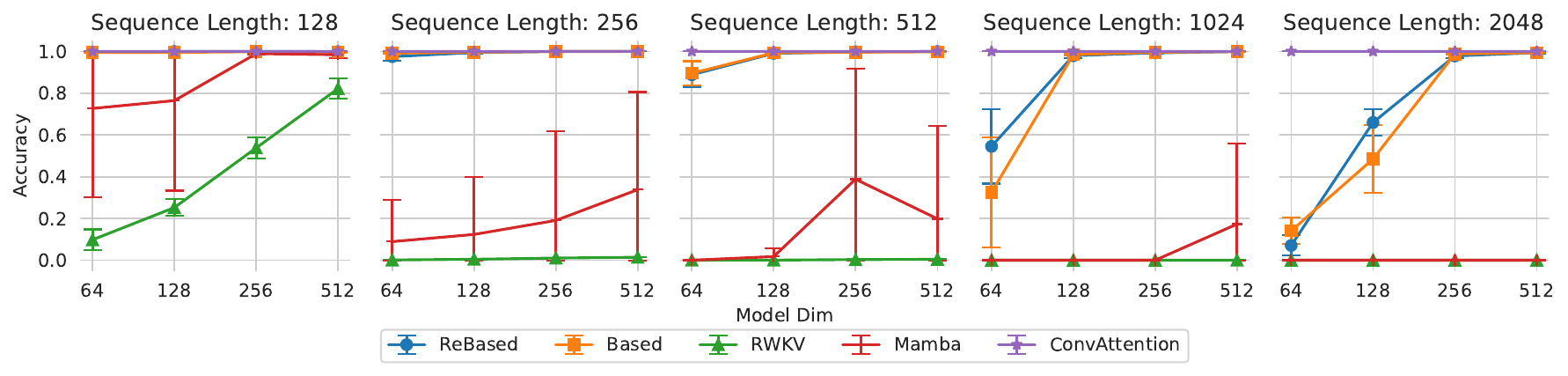}
    \caption{Results on the MQAR dataset, designed to measure In-Context Learning capabilities of an architecture \citet{mqar}. ReBased outperforms all baselines except Attention across different sequence lengths and model sizes. See Section \ref{sec:mqar} for more details.}
    \label{fig:main}
\end{figure*}

A profound comprehension of the processes occurring within the Based architectures is essential for their advancement. Upon examining how attention scores are distributed, we argue that the kernel function previously adopted in Based cannot be considered optimal, resulting in limitations when dealing with lengthy context and small model capacity. To address this issue, we introduce \textbf{ReBased} (Revisited Based), a novel variation of the Linear Transformer model that improves the use of attention kernels. The crux of our development lies in addressing the inability of Based to disregard specific tokens with zero probability during the attention process. By refining the kernel function and incorporating new architectural modifications, we have created a model that improves accuracy on tasks involving retrieving information from long sequences of tokens while simplifying the calculation of the attention mechanism.

When testing our enhanced architecture on the MQAR task, we found that ReBased surpasses the original Based model across a variety of contexts and model sizes. Additionally, after training with the Pile dataset \citep{pile}, we observed that ReBased performs better than its predecessor at In-Context Learning and excels at modeling associative dependencies measured through improved perplexity metrics.

\section{Recent Work}
The Vanilla Transformer architecture \citep{transformer}, although widely used in NLP \citep{gpt2, llama2, bert, mistral}, suffers from growing computational and memory demands ($\mathcal{O}(d*N^2)$ as sequence lengths ($N$) and head size ($d$) increase). While this is not much of a problem when it comes to shorter sequences, it becomes a significant bottleneck when working with longer ones.

Several alternative architectures were proposed to address this issue. \citet{linear} suggested substituting the attention mechanism's exponential function, which is meant to measure the similarity between queries and keys, with a product of kernel functions that can be separately evaluated for queries and keys. This kernel-based approach reshapes the computation within the attention mechanism, cutting the time and memory complexity to $\mathcal{O}(d^2*N)$. Additionally, during inference, it supports sampling sequences with linear complexity regarding length, similar to RNNs \citep{lstm, gru}.

In a different approach, State Space Models (SSMs) borrow from control theory to offer a simplified structure akin to RNNs, but without activation functions across time \citep{s4, s5, hippo}. The Mamba model, also known as S6 \citep{mamba}, stands out in this category, displaying enhanced learning of short-term dependencies in texts compared to existing pre-trained LLMs \citep{mistral, llama2}.

Despite these advancements, there is no standard way to fully evaluate these innovative architectures to assess their performance limits. One standard evaluation method is to pre-train a language model and assess its perplexity with a given dataset, but this may not truly reflect the model's ability to manage long context dependencies. Another option is to use the Long Range Arena (LRA) benchmark, which involves classification tasks with long input sequences. Though some new models have outperformed Transformers in the LRA, it is believed that the benchmark is capable of introducing bias in the comparison \citep{scratch}.

One promising evaluation approach is to test an architecture's In-Context Learning abilities. \citet{ar} introduced the concept of Associative Recall (AR), a task where the model learns to copy a token from a sequence after a certain point. However, while in \citet{h3} the associative recall task was implemented with a goal to retrieve only one token, \citet{mqar} noted that this task could be considered overly simplistic. This led to the creation of the Multi-Query Associative Recall (MQAR) task, which requires retrieving multiple tokens from the context. Findings on MQAR indicate that while newer models may compete with Transformer in terms of perplexity, they can still struggle with long contexts at small model sizes because of their limited In-Context Learning capabilities. Meanwhile, Transformers remain robust against such factors. Lastly, \citet{based} introduced Linear Transformer with a new kernel function (namely Based), showcasing enhanced performance on the MQAR task when compared to Mamba. 

Despite this improvement, compared to traditional Transformers, the problem of decline in performance when handling long sequences with smaller models still remains. Addressing this challenge is the primary goal of our paper.

\section{Background}
\subsection{Linear Transformers}

To fully grasp the Based architecture, it is vital to first  discuss the original Transformer model. The attention mechanism, which is central to the Transformer's functionality, evaluates the output $y_i$ for each position $i$ as follows

\begin{equation*}
  y_i = \frac{\sum_{j=0}^i \text{sim}(Q_i, K_j)V_j}{\sum_{n=0}^i\text{sim}(Q_i, K_n)},
\end{equation*}

where the term $\text{sim}(Q_i, K_j) = \exp\left(\frac{Q^T_i K_j}{\sqrt{d}}\right)$ represents the similarity between the query $Q_i$ and the key $K_j$ using an exponential function. Despite its effectiveness, the original Transformer's reliance on this attention mechanism incurs a quadratic increase in both computational time and memory use as the sequence length grows, which becomes impractical for processing long sequences.

To address this scalability problem, \citet{linear} suggested replacing the direct computation of similarity between $Q$ and $K$ with a transformation through a non-linear kernel function $\phi(\cdot)$. This allows for the following approximation: $\text{sim}(Q_i, K_j) \approx \phi^T(Q_i)\phi(K_j)$. By implementing this kernel, the Linear Transformer computes $y_i$ as

\begin{equation*}
  y_i = \frac{\sum_{j=0}^i \phi^T(Q_i)\phi(K_j) V_j}{\sum_{n=0}^i\phi(Q_i)\phi^T(K_n)}.
\end{equation*}

By rearranging the operations, we can express the computation as

\begin{equation*}
  y_i = \frac{\phi^T(Q_i)\sum^i_{j=0}\phi(K_j)V^T_j}{\phi^T(Q_i)\sum^i_{n=0}\phi(K_n)}.
\end{equation*}

By calculating $\phi(K_j)V^T_j \in \mathbb{R}^{d \times d}$ upfront, the complexity of the attention mechanism transitions to linear with the sequence length, addressing the inefficiencies of the original model.

\subsection{Based}

Selecting an appropriate kernel function $\phi(\cdot)$ is critical to a Linear Transformer's performance. Various kernel functions have been proposed \citep{random_feature, fast_weight, cos_former}, but on language modeling tasks, none have surpassed the original attention mechanism. However, a breakthrough was achieved by \citet{based}, who introduced a novel kernel function inspired by the Taylor series expansion of the exponential function, defined as

\begin{equation*}
  \text{sim}(q, k)= 1 + q^Tk + \frac{(q^Tk)^2}{2}.
\end{equation*}

The choice of this kernel is motivated by its ability to approximate the exponential function over a specific range of $q^Tk$ values. In addition, \citet{based} utilized a hybrid architecture by combining linear attention with convolutional layers since doing so was shown to help models handle short non-associative dependencies in the sequences \citep{h3, poli2023hyena, fu2023simple}

 In doing so, when evaluated on the MQAR task, the Based model demonstrated that it was capable of outperforming the Mamba model \citep{mamba} under circumstances of substantial context length and constrained model capacity due to smaller sizes. Nevertheless, compared to the original Transformer, a discernible drop-off in performance remains, indicating room for further improvement.

\section{Revisiting Based}
\label{sec:lin_trans}

\begin{figure}
    \centering
    \includegraphics[width=0.8\linewidth]{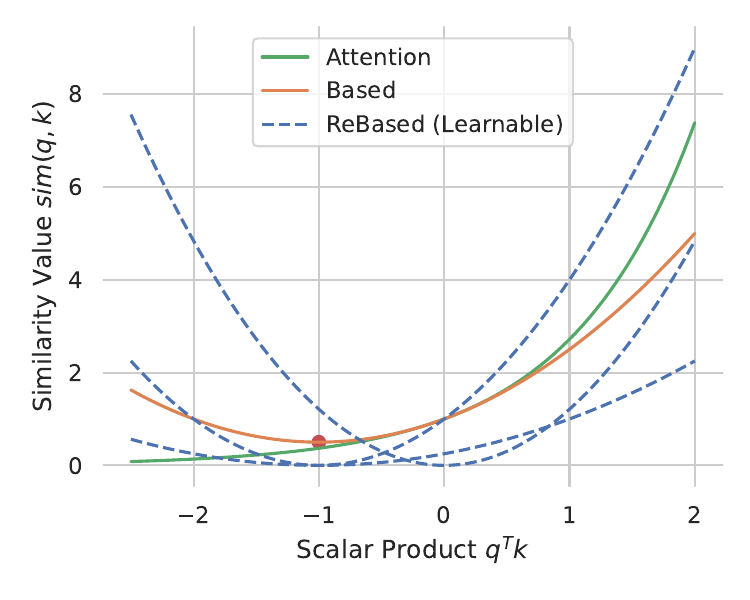}
    \caption{Similarity between $q$ and $k$ with respect to scalar product. Note that the Based model has a minimal $\text{sim}(q, k)$ value of 0.5, which can lead to suboptimal performance. We propose to learn the scale and shift of the parabola jointly with the model and make it possible to zero out the similarity value. See Section \ref{sec:lin_trans} for more details  and Section \ref{section:setup} for experimental setup description.}
    \label{fig:sim}
\end{figure}

In our study, we explore the fundamental requirements for kernel functions. We examine the exponential function and its approximate representation, as depicted in Figure \ref{fig:sim}. We observe a limitation in the approximation since its minimal value is fixed at $0.5$. This is problematic for handling long sequences, as it is difficult to assign a near-zero attention score to specific token pairs. Ideally, we want to be able to diminish the attention scores to zero, which would require significantly larger values elsewhere in the normalization process with the Based model.

To rectify this issue, a straightforward approach would be to adjust the lowest point of the kernel function to zero. However, this solution prompts us to ask why the minimum value of the kernel function should occur precisely at $q^Tk = -1$. As used in the original Transformer, the traditional exponential similarity function increases monotonically, but the quadratic kernel has an optimal value to which it decreases and then ascends from. Therefore, to decrease attention in the Transformer, one would aim to minimize $q^Tk$. In contrast, the ideal $q^Tk$ should be exactly $-1$ for the Based method. Otherwise, the attention score would increase. This condition may induce less-than-ideal training outcomes and degrade the model's accuracy.

These challenges lead us to conjecture that if the quadratic kernel is used to calculate the similarity between $q$ and $k$, we must consider the range of potential $q^Tk$ values and create adjustable parameters for the parabolic function to align with these values during training. Simplifying for clarity, let us look at a one-dimensional scenario. We can express the trainable parameters of the kernel function in relation to the affine transformation of $q$ and $k$ as such

\begin{align*}
  q^\prime = &\gamma_Q \cdot q + \beta_Q,\ \ k^\prime = \gamma_K \cdot k + \beta_K; \\
  &\text{sim}(q^\prime, k^\prime) = \phi^T(q^\prime) \phi(k^\prime).
\end{align*}

Here, $\phi(\cdot)$ represents a quadratic function. The model can learn any quadratic function with a determined minimum value by adjusting its parameters. We can, therefore, simplify the kernel function to

\begin{equation*}
  \phi(x) = {x}^2.
\end{equation*}

Incorporating the affine transformation into the kernel function, we obtain

\begin{align*}
  \phi(q^\prime) & = (\gamma_Q \cdot q + \beta_Q)^2 = \gamma_Q^2 q^2 + 2\gamma_Q\beta_Q q + \beta_Q^2, \\
    \phi(k^\prime) & = (\gamma_K \cdot k + \beta_K)^2 = \gamma_K^2 k^2 + 2\gamma_K\beta_K k + \beta_K^2.
\end{align*}

where $q$ and $k$ have their unique parameters $\gamma_Q$, $\gamma_K$, $\beta_Q$, and $\beta_K$, enabling the model to learn any quadratic function that is non-negative and has a single real root.

Interestingly, our transformation resembles the application of Layer Normalization \citep{ln}, minus the normalization itself. We hypothesize whether normalizing $q$ and $k$ before the kernel function could improve the model's performance. Our suspicion is confirmed when normalization enhances results, as demonstrated in a later Ablation study. Consequently, our refined \textit{ReBased} model incorporates Layer Normalization.

\begin{figure*}[!hbtp]
    \centering
    \includegraphics[width=\linewidth]{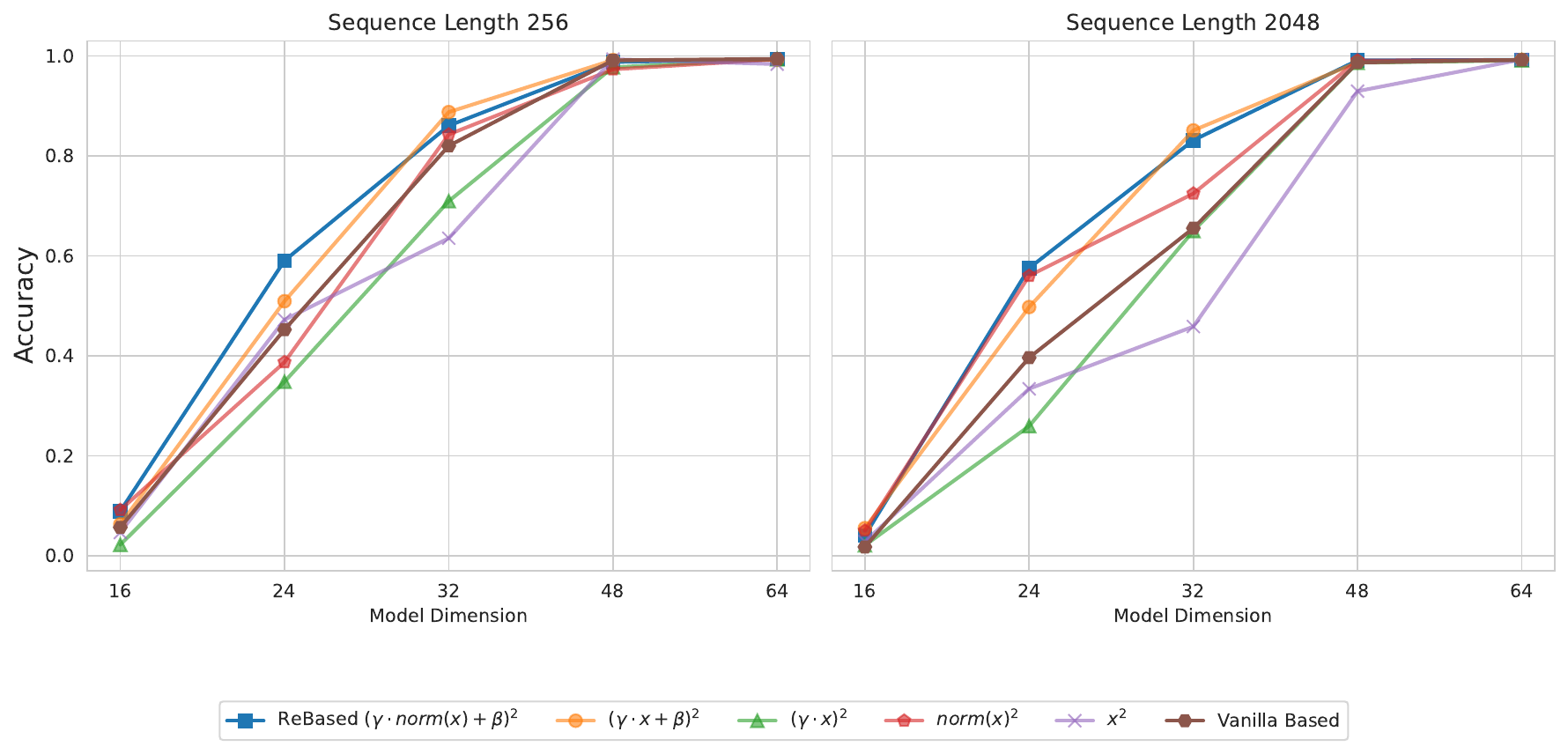}
    \caption{Ablation study for the proposed modifications. For sequence length $256$, the difference is not very significant. Nevertheless, the ReBased model performs best on all model dimensions. With a sequence length of $2048$, the difference becomes more evident. Unlike Based, the ReBased model retains performance across long and short sequences. See Section \ref{sec:ablations} for the experiment setup and extended description of our results and Section \ref{section:setup} for experimental setup description.}
    \label{fig:ablations}
\end{figure*}

In the following sections, we provide an in-depth analysis and conduct comprehensive experiments to validate the effectiveness of these modifications.

\section{Experiments}

\subsection{Experimental Setup}
\label{section:setup}

We applied the first evaluation of our ReBased model on the \textbf{MQAR task}, for which we trained a model to perform associative recall with varying numbers of retrieved tokens. \citet{mqar} suggested that for a comprehensive assessment, models need to be tested across different sequence lengths, model sizes, and number of query-key pairs to be retrieved. However, those experiments were limited, only exploring sequence lengths up to $512$. These constraints resulted in the Based model displaying performance comparable to the traditional attention mechanism.

Longer sequence lengths can be explored to gain a deeper understanding of how improvements in associative recall are affected by changes in model configurations. This is why we extended our training to include models capable of handling sequence lengths of $[128, 256, 512, 1024, 2048]$. We tested a range of hidden sizes from $64$ to $512$. For our ablation study to yield more precise insights, we also employed smaller models with hidden sizes as modest as $[16, 24, 32, 48]$.

In order to tailor our approach to varied sequences, we used different query-key (qk) pairs for each length. The specifics of these configurations are detailed in Appendix \ref{ap:mqar_details}.

We also put other sub-quadratic architectures to the test, including Mamba (SSM family) \citep{mamba}, Hyena (the long convolutions family) \citep{poli2023hyena}, the vanilla attention method, and RWKV \citep{rwkv}. By comparing a diverse range of models, our goal was to present a well-rounded evaluation of how our ReBased model stands out in the field. For Based, we utilized Triton kernels published by \citet{triton_kernels}, and for ReBased, we modified it so that $\phi(x) = x^2$.

We used a hybrid architecture with short convolution and kernel size 3 in the first layer, and specified a mixer in the second. We found that this setup was more stable on longer sequence lengths, especially when using an attention mixer. However, we did not modify the Mamba model since convolutions were already present inside the Mamba block. We put the full results and model architecture details in Appendix \ref{ap:mqar_details}.

\begin{table*}[!ht]
    \centering
    \scalebox{0.68}{
    \begin{tabular}{r|cccc|c|cccc|c} 
\toprule & \multicolumn{4}{|c}{Sequence Length 256}& & \multicolumn{4}{|c}{Sequence Length 2048} \\ 
Architecture & $16$ & $24$ & $32$ & $48$ & Mean & $16$ & $24$ & $32$ & $48$ & Mean \\ \midrule
Based & $0.06 \pm 0.02$ & $0.45 \pm 0.15$ & $0.82 \pm 0.06$ & $0.99 \pm 0.00$& 0.58  & $0.02 \pm 0.02$ & $0.40 \pm 0.18$ & $0.66 \pm 0.08$ & $0.99 \pm 0.01$ & 0.51\\ 
$x^2$ & $0.05 \pm 0.05$ & $0.47 \pm 0.08$ & $0.64 \pm 0.42$ & $0.99 \pm 0.00$ & 0.54 & $0.03 \pm 0.03$ & $0.33 \pm 0.21$ & $0.46 \pm 0.42$ & $0.93 \pm 0.09$ & 0.44\\ 
$norm(x)^2$ & $0.09 \pm 0.04$ & $0.39 \pm 0.24$ & $0.84 \pm 0.09$ & $0.97 \pm 0.02$ & 0.57 & $0.05 \pm 0.05$ & $0.56 \pm 0.10$ & $0.72 \pm 0.17$ & $0.99 \pm 0.00$ & 0.58\\ 
$(\gamma \cdot x)^2$ & $0.02 \pm 0.02$ & $0.35 \pm 0.22$ & $0.71 \pm 0.09$ & $0.98 \pm 0.03$ & 0.51 & $0.02 \pm 0.03$ & $0.26 \pm 0.45$ & $0.65 \pm 0.37$ & $0.99 \pm 0.01$ & 0.48\\ 
$(\gamma \cdot x + \beta)^2$ & $0.06 \pm 0.01$ & $0.51 \pm 0.08$ & $0.89 \pm 0.03$ & $0.99 \pm 0.00$ & 0.61 & $0.06 \pm 0.03$ & $0.50 \pm 0.08$ & $0.85 \pm 0.04$ & $0.99 \pm 0.01$ & 0.60\\ \midrule
\specialcell[c]{ReBased \\ \tiny{$(\gamma \cdot norm(x) + \beta)^2$}} & $0.09 \pm 0.05$ & $0.59 \pm 0.06$ & $0.86 \pm 0.08$ & $0.99 \pm 0.00$ & \textbf{0.63} & $0.04 \pm 0.03$ & $0.58 \pm 0.01$ & $0.83 \pm 0.04$ & $0.99 \pm 0.00$ & \textbf{0.61}\\ 
\bottomrule
    \end{tabular}}
    \caption{Ablation study for proposed modifications with standard deviation across $5$ seeds. See Figure \ref{fig:ablations} for a visual presentation of the results, Section \ref{sec:ablations} for experiment setup and extended result description and Section \ref{section:setup} for the description of our experimental setup.}
    \label{tab:ablation}
\end{table*}

In \textbf{language modeling}, our second experimental setup leveraged the extensive Pile dataset \citep{pile} to train a language model (LM). We opted for a sequence length of 4096, a slight increase from the standard value while still ensuring the replication of the architectural framework as presented by \citet{based}\footnote{The experiment details can be found in a \href{https://hazyresearch.stanford.edu/blog/2023-12-11-zoology2-based}{blog post} and \href{https://wandb.ai/hazy-research/zg-neox/reports/Zoology---Vmlldzo2MjExMTc2?accessToken=z1e7bu20nm8r5l8eqp2e9ftork7o94tthd2nq6wimai1k4ger1f3vhb76rfvliw2}{WandB report} associated with the main paper.}. Note that some hyperparameters such as model dimension and the number of layers were set in order to match the number of model parameters in the initial experiment. Detailed model configuration can be found in Appendix \ref{ap:lm}. 

The MQAR task provided insights into In-Context Learning proficiencies across various architectures, while the language modeling assessment allowed us to appraise short-term dependency modeling capacities. Beyond traditional perplexity metrics on validation data, we also scrutinized the Associative (AR) and Non-Associative (Non-AR) variants of perplexity. Here, AR corresponds to token positions necessitating associative recall, while Non-AR refers to other tokens. When tokens recur within a text, the subsequent appearances are categorized as AR, highlighting the model’s capability to recall from context.

\subsection{MQAR experiment}
\label{sec:mqar}

In Figure \ref{fig:main}, we present the capability of various models to handle the MQAR task as the sequence length increases. One key observation is that, at a sequence length of 2048, all models, except for the Attention model, struggled to perform effectively when limited to a model dimension of 64. As we expanded the model dimensions, the performance of the ReBased model matched or surpassed the Based model. The RWKV and Mamba architectures failed on the MQAR task across all tested model sizes.
 
This experiment highlights the significance of utilizing more sophisticated setups, as the performance discrepancy between the Attention model and the other models (Based and ReBased) becomes pronounced only when the sequence length exceeds 512. These results suggest that the efficacy of attention alternatives like ReBased becomes particularly important when processing long sequences. Therefore, more consideration should be devoted to configurations involving lengthy sequences to leverage the full potential of such models.

\subsection{Ablation Study}
\label{sec:ablations}

We comprehensively examined the individual elements of our ReBased model to understand how each of them contributes to its overall effectiveness, and ensure the transparency of our findings. Our experiments were meticulously designed to evaluate the model by assessing the influence of its separate components on performance. The experimental configurations were as follows:

\begin{itemize}
    \item $x^2$ -- substituting the original kernel function with a simple element-wise squaring operation, $\phi(x) = x^2$.
    \item $norm(x)^2$ -- integrating a normalization step without an affine transformation before applying the squaring operation.
    \item $(\gamma\cdot x)^2$ -- introducing an affine transformation solely in terms of scaling (without bias) for the queries and keys. 
    \item $(\gamma\cdot x + \beta)^2$ -- incorporating affine transformation with both scaling and bias for the queries and keys. 
    \item ReBased $(\gamma\cdot norm(x) + \beta)^2$ -- our comprehensive model, which involves normalization and affine transformation, including bias, for queries and keys.
\end{itemize}

% \begin{figure*}[!ht]
%     \centering
%     \includegraphics[width=\linewidth]{pics/ablations_evp.pdf}
%     \caption{Expected validation accuracy across different hyperparameters. The ReBased model works best across all hyperparameters, budgets, and model dimensions. See Section \ref{sec:ablations} for more details.}
%     \label{fig:evp_ablations}
% \end{figure*}

Note that for $q$ and $k$, there are different scaling parameters $\gamma_Q$, $\beta_Q$, $\gamma_K$, and $\beta_K$ for each experiment involving affine transformation.

Our goal is to highlight the effect of sequence length variability in the MQAR task on model performance. For this evaluation, we standardized the number of retrieval pairs to 32. Theoretically, no impact on performance should be observed, as the amount of information required to be stored in the hidden states is sequence-length agnostic. We investigated the effects on sequences of lengths 256 and 2048 and illustrated our findings in Figure \ref{fig:ablations} (also available in Table \ref{tab:ablation} with a standard deviation of accuracy across $5$ seeds). We must emphasize the significance of long context setups evaluated in our experiments. Its characteristics are vital, as successful performance on long sequences highlights the capability of the model to make full use of its architectural innovations. It also translates into notable practical advantages in real-world applications where handling extensive context efficiently can be crucial.

The proposed ReBased model performs better than every other modification. Performance on the short $256$ length is less noticeable than on the long $2048$ sequence length. We see a performance drop from simply replacing the original kernel function with $x^2$. We presume that this is caused by suboptimal scale of features, since by placing normalization before the kernel function, we can notice a performance increase even in comparison to the Based model. Affine transformations $(\gamma \cdot x)^2$ and $(\gamma \cdot x + \beta)^2$ also show favorable performance compared to the $x^2$ model, which does not significantly decrease with sequence length.

% In our experiments, we found ReBased to be more stable during training with various hyperparameters. To demonstrate this, we utilize an Expected Validation Performance (EVP) plot \citep{evp}. We treat the average across five seeds as the final accuracy. Our results are presented in Appendix Figure \ref{fig:evp_ablations}. We noticed that even in cases where the model dimension is sufficiently large to store all necessary information, our modifications lead to $100\%$ accuracy for every hyperparameter set and every seed we use, in contrast with the Based model, where we observe degradation for certain learning rates.

\subsection{Language Modeling}
\label{sec:lm}

\begin{table}[!h]
    \centering
    \scalebox{1.}{
    \begin{tabular}{r|ccc} \toprule
      & \multicolumn{3}{c}{Perplexity} \\
     Architecture & All & AR & Non-AR \\ \midrule
     % Attention & 16.62 & \textbf{3.57} & 58.57 \\
     Attention & 11.98 & 3.07 & 33.95 \\ \midrule
     Based    & 12.99 & 3.27 & 37.02 \\
      ReBased  &  \textbf{12.90} & \textbf{3.25} & \textbf{36.73} \\ \bottomrule
    \end{tabular}}
    \caption{Perplexity results on Pile \cite{pile} dataset. ReBased improves the result on AR tokens. However, there is still a small gap between Attention and ReBased. See Section \ref{sec:lm} for more details and Section \ref{section:setup} for experimental setup description.}
    \label{tab:perplexity}
\end{table}

We conducted experiments with language modeling following the setup described in Section \ref{section:setup}. See Table \ref{tab:perplexity} for the results.

We note that ReBased model performs better than Based on both AR and non-AR tokens leading to lower overall perplexity. This can be considered as a sign of better  In-Context Learning performance. In the next section we consider few-shot setup on several tasks to validate 

When considering AR perplexity, we observe that there is still a gap between the vanilla Transformer architecture and alternative models, which is aligned with the results on the MQAR dataset. However, we note that ReBased still performed better than Based. Regarding Non-AR perplexity, ReBased outperformed both Based architectures, leading to better overall perplexity value. Note that attention has slightly more trainable parameters, see Appendix \ref{ap:lm} for more details.

These results suggest that, despite language modeling perplexity being lower for an alternative to Transformer architectures \citep{based, mamba}, this may be achieved due to better short-term dependency modeling, which does not require learning associative operations necessary to perform In-Context Learning \citep{ar}. The vanilla Transformer still performs best in terms of its ability to attend to some token in-context.

\subsection{Few-Shot Performance}
\label{sec:fs}

To further explore the ability of the model to improve results on real-world scenarios we validate trained Based and ReBased models on a common few-shot benchmarks from the LM Evaluation Harness \cite{eval-harness} benchmark and SuperGLUE \cite{superglue}. Results are presented in Table \ref{tab:lm_eval} and Table \ref{tab:superglue}. ReBased outperforms Based on the most of the tasks.

\begin{table*}[!h]
    \centering
    \scalebox{0.9}{
    \begin{tabular}{r|cccccc} \toprule
    Architecture & Winogrande & Piqa & Hellaswag & Arc-E & Arc-C & Macro\\ \midrule
         Based &  50.4 & 62.1 & \textbf{30.8} & 40.4 & \textbf{22.9} & 41.3 \\
         ReBased & \textbf{54.6} & \textbf{62.8} & 30.7 & \textbf{41.0} & 21.5 & \textbf{42.1} \\\bottomrule
    \end{tabular}}
    \caption{1-shot performance on tasks from LM evaluation harness benchmark \cite{eval-harness}. See Section \ref{sec:fs} for more details.}
    \label{tab:lm_eval}
\end{table*}

\begin{table*}[!h]
    \centering
    \scalebox{0.9}{
    \begin{tabular}{r|ccccccc} \toprule
    Architecture & WSC & WIC & RTE & Record (F1/EM) & MultiRC & Copa & BoolQ\\ \midrule
         Based &  55.8 & 46.5 & 47.6 & 62.7/62.1 & 51.5 & 66.0 & 48.3 \\
         ReBased & \textbf{56.7} & \textbf{46.9} & \textbf{53.1} & \textbf{62.8/62.2} & \textbf{51.9} & \textbf{67.0} & \textbf{52.0} \\\bottomrule
    \end{tabular}}
    \caption{1-shot performance on SuperGLUE benchmark \cite{superglue}. See Section \ref{sec:fs} for more details. }
    \label{tab:superglue}
\end{table*}

\subsection{Analysis}
\label{sec:analysis}

\begin{figure*}[!htbp]
    \centering
    \includegraphics[width=0.8\linewidth]{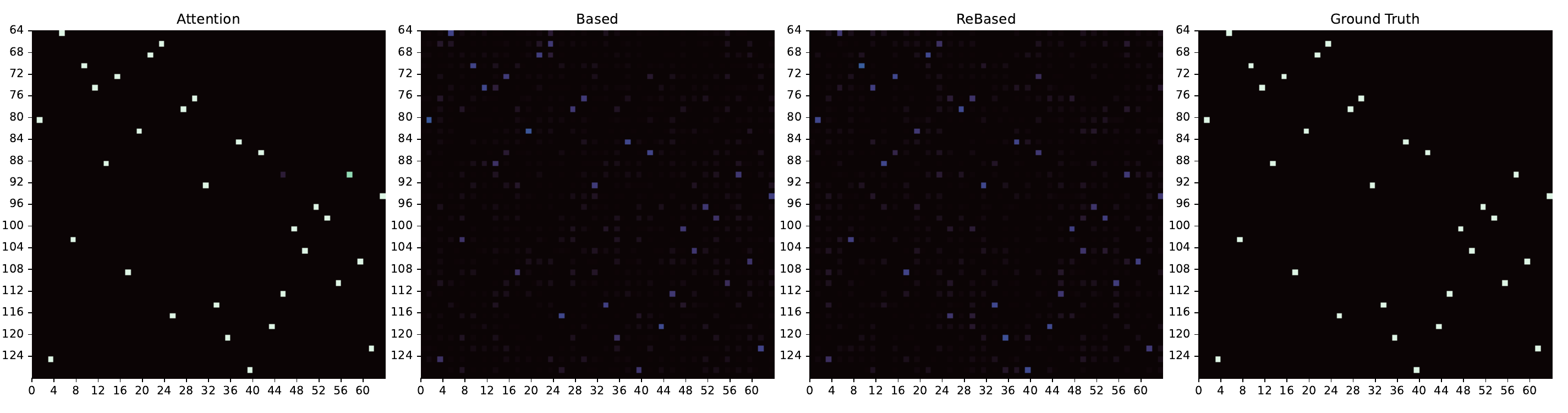}
    \caption{Attention matrix for the different models, and ground truth positions for the query. We measure IoU between model's attention and ground truth matrix for $10000$ examples. Illustration of the experiment is described in Section \ref{sec:analysis} Results are presented in Table \ref{tab:iou}.}
    \label{fig:iou}
\end{figure*}

In this section, we delve into the internal dynamics of the ReBased model by examining attention matrices, which are commonly used to elucidate the decision-making of models and the flow of information between tokens. Notably, we can use the parallel mode with both Based and ReBased models to construct these matrices.

For our analysis, we employ the MQAR dataset \cite{mqar} and train a compact model configured with a sequence length of 128 and 32 retrieval pairs. To ensure clear interpretation of the attention maps, we used fixed weights in the first layer, which consists of a short convolution with a kernel that attends to the previous token. Following the training phase, we compute the Intersection over the Union (IoU) metric between the attention matrix and the actual positions of the tokens that are to be retrieved. The correct positions are crucial, as they represent the locations from which the model must copy the hidden states in order to successfully resolve the task. This copying mechanism is particularly vital and is implemented via focused attention in the second layer of the network \cite{ar}. Consequently, the IoU provides a quantitative measure of how well our model has learned to replicate this crucial pattern of token retrieval. A visualization of this phenomenon using IoU on a randomly selected example from the dataset is shown in Figure \ref{fig:iou}. Note that we cropped attention matrix to examine only a region where qk-pairs stored.

\begin{table}[!h]
    \centering
    \scalebox{0.9}{
    \begin{tabular}{r|c|c} \toprule
    Architecture & IoU & Accuracy\\ \midrule
        Attention &  \textbf{0.999} & 1\\
         Based &  $0.157$ & 0.956\\
         ReBased & 
         \textit{0.173} & 0.957 \\\bottomrule
    \end{tabular}}
    \caption{IoU with attention matrix and ground truth position to retrieve on the MQAR task for $10000$ examples. Detailed experiment setup can be found in \ref{sec:analysis}. }
    \label{tab:iou}
\end{table}

Our results are presented in Table \ref{tab:iou}. In our experiment, the Attention model yielded a superior IoU score compared to both the Based and ReBased models. However, the ReBased model shows promise in narrowing the performance divide that exists between sub-quadratic methods and the attention-based model. This suggests that, despite the relative simplicity of the method, it could serve as an informative metric for the MQAR dataset, particularly when the accuracy score is close to one, making it challenging to discern the performance differences between models in more intricate testing scenarios.

\section{Conclusion and Future Work}

In this paper, we present ReBased, a novel architecture for sub-quadratic attention computation. For our model, we analyzed the Base architecture and proposed to develop it even further by using polynomial kernels with learnable parameters and adding normalization before the kernel evaluation. While incorporating layer normalization into model training was attempted previously \citep{querykey}, our method integrates this normalization directly into the kernel function. With this simple architectural change, we achieved results that outperformed Based on MQAR and language modeling with the Pile dataset tasks. 

We analyzed the internal representations of ReBased, Based, and vanilla attention modules, and concluded that ReBased resembles attention more than Based. Notably, while Based uses a Taylor expansion of an exponential function, a ReBased kernel function is different from the exponent but shows better performance. Our research suggests that using a second-order polynomial might be insufficient for the best performance, and indicates that more sophisticated learnable kernels could be utilized to improve the performance of trained models. Normalization could further improve various kernel functions. This highlights a need for researchers to revisit kernel-based methods with the goal of enhancing their adaptability and efficiency. 

Our findings reveal a disparity in handling the MQAR task between attention-based models and others such as Based, specifically as sequence lengths increase. Attention models excel on longer sequences, significantly outperforming their non-attention counterparts. These results highlight the necessity of further research into strategies that could bridge this gap in order to reach the performance of attention-based methods. Perhaps the superior aspects of attention mechanisms could be matched or exceeded by other models, especially on tasks that require associative recall, such as machine translation \cite{ssm_nmt}. Future research could give insight into this, leading to improved models for processing long sequences.

\section{Limitations}

While our proposed method demonstrates applicability to a wide range of tasks typically addressed by Transformers, its effectiveness in handling tasks involving intensive copying or recalling previous context remains unclear (see Table \ref{tab:perplexity} and \citet{jelassi2024repeat}). Successfully addressing these tasks is crucial for fully mitigating inference problems associated with attention mechanisms.

It is also worth noting that our experiments are limited to academic-scale models. This does pose certain limitations, particularly in extrapolating the findings to larger models. However, given the resource constraints, our results still provide valuable insights into the potential efficacy of our method.

%In our work we proposed a new method called ReBased, which surpasses all popular linear-time baselines in terms of MQAR task and beats Based model on a Language Modeling task. We delved into the attention map differences between our method, Attention and Based, and found clear explanation of the performance gap even when accuracy on MQAR were equal to 1. We believe that trainable composition of affine transforms and kernel function could be beneficial for the others kernel functions to be explored and further improve training procedure to be more robust to hyper-parameters, which become crucial on a larger scales. We show that ReBased method is closer to attention in terms of attention maps, despite disentangle kernel function from Taylor's series of exponent. We also demonstrate that there is still a gap to be diminished.
\bibliography{custom}

\appendix
\label{sec:appendix}

\begin{figure*}[t!]
  \centering
  \begin{subfigure}[b]{0.5\textwidth}
    \centering
    \includegraphics[width=\linewidth]{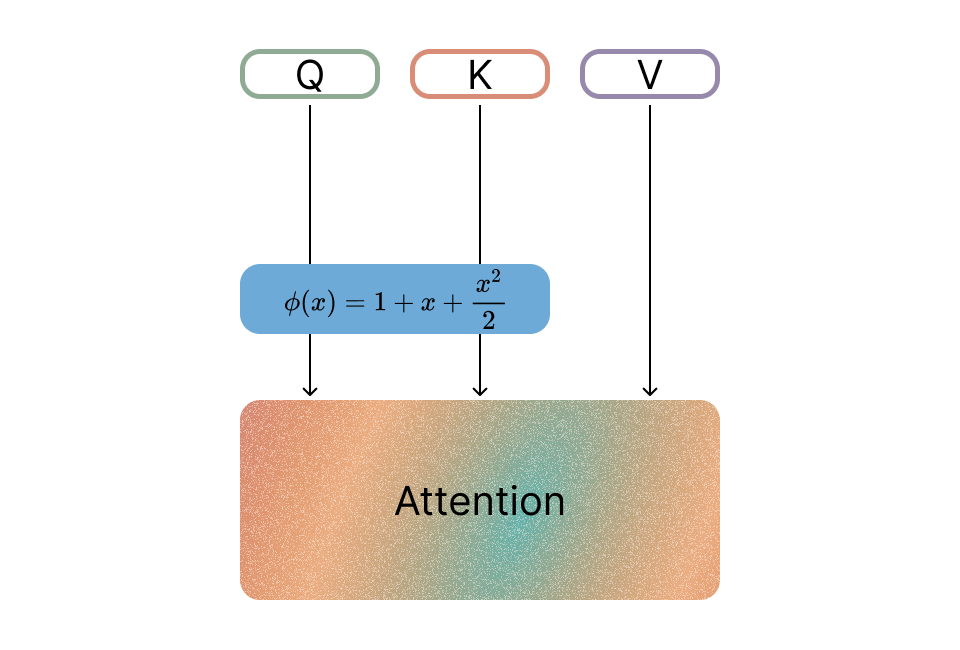}
    \caption{Based architecture}
    \label{fig:subfig1}
  \end{subfigure}%
  \hfill
  \begin{subfigure}[b]{0.5\textwidth}
    \centering
    \includegraphics[width=\linewidth]{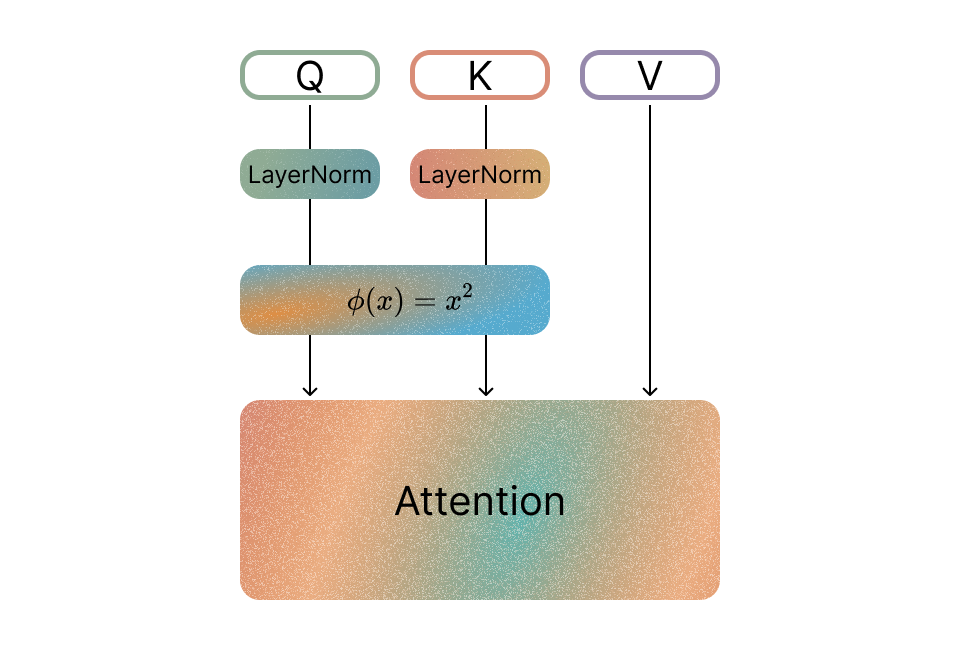}
    \caption{ReBased architecture.}
    \label{fig:subfig2}
  \end{subfigure}
  \caption{Architectures visualization.}
  \label{fig:subfigures}
\end{figure*}

\begin{table*}[!hbtp]
    \centering
    \scalebox{0.8}{
    \begin{tabular}{c|cccccc}
\toprule Model Dimension & Attention & ConvAttention & RWKV & ConvRWKV & Mamba & Based (Rebased) \\ \midrule

 64 & 623744 & 578752 & 623872 & 677120 & 655360 & 577984 $(+768)$\\
 128 & 1313024 & 1149312 & 1313280 & 1395200 & 1413120 & 1179520 $(+768)$ \\
 256 & 2888192 & 2462464 & 2888704 & 2561024 & 3235840 & 2459392 $(+768)$\\
 512 & 6824960 & 5580288 & 6825984 & 5777408 & 7847936 & 5307904 $(+768)$ \\ \bottomrule
\end{tabular}}
    \caption{Number of model parameters in MQAR dataset. See Appendix \ref{ap:mqar_details}.}
    \label{tab:parameter_count}
\end{table*}

\section{Details for the MQAR dataset experiments}
\label{ap:mqar_details}

\begin{figure}[!ht]
    \centering
    \includegraphics[width=\linewidth]{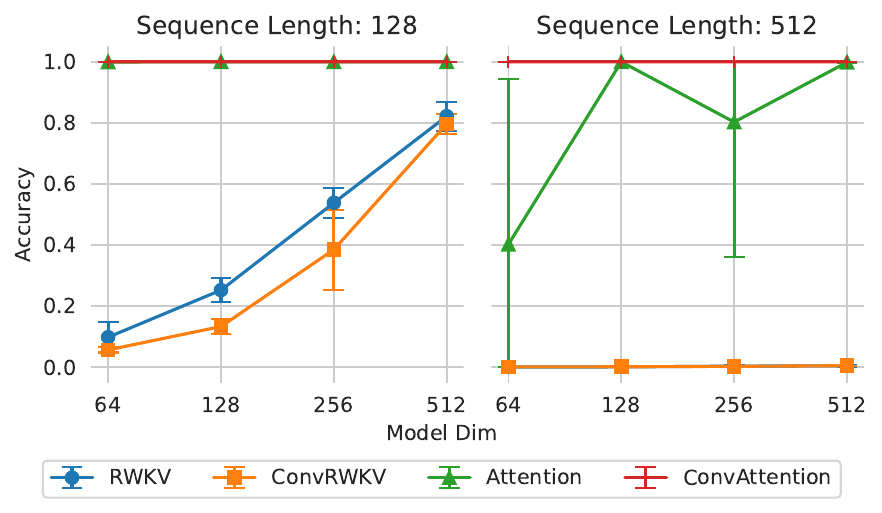}
    \caption{Performance of the hybrid architecture with convolutions on the first layer and the vanilla architecture.}
    \label{fig:ap_fig}
\end{figure}

In our experiments, we use the code from the official MQAR repository \citep{based}\footnote{\href{https://github.com/HazyResearch/zoology}{https://github.com/HazyResearch/zoology}}. However, we modify the attention model from the one reported in \citet{based}, as we found it more stable (see Figure \ref{fig:ap_fig}). We can see that replacing the first attention layer is beneficial for performance. RWKV performs better when we do not replace the first layer, which is why we use we use two RWKV layers in our main experiment (see Figure \ref{fig:main}). We report the number of trainable parameters in Table \ref{tab:parameter_count}.

We also modify the data configs to be more challenging for the model. You can see the adjusted parameters in Table \ref{tab:qk}.

\begin{table}[!h]
    \centering
    \begin{tabular}{|c|c|} \toprule
    seq\_length & qk\_pairs\\ \midrule
        128 &  16\\
        256 &  64\\
        512 &  128\\
        1024 &  256\\
        2048 &  512\\ \bottomrule
    \end{tabular}
    \caption{Sequence lengths and number of QK pairs in dataset.}
    \label{tab:qk}
\end{table}

We use a batch size of 512 for all experiments. In cases where there is not enough GPU memory, we use the gradient accumulation technique. For the learning rate, we use hyperparameter search with the following grid: 5e-4, 1e-3, 3e-3, 1e-2. We use five different seeds for all reported results.

\section{Stability}

\begin{figure*}[!p]
    \centering
    \includegraphics[width=\linewidth]{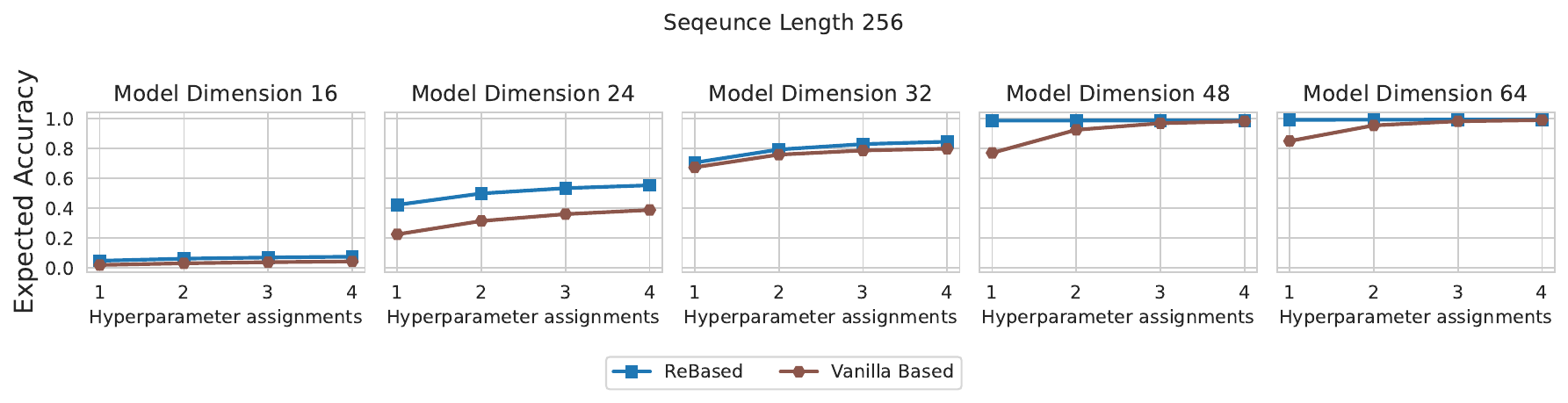}
    \caption{Expected validation accuracy across different hyperparameters. The ReBased model works best across all hyperparameters, budgets, and model dimensions. See Section \ref{sec:ablations} for more details.}
    \label{fig:evp_ablations}
\end{figure*}

In our experiments, we found ReBased to be more stable during training with various hyperparameters. To demonstrate this, we utilize an Expected Validation Performance (EVP) plot \citep{evp}. We treat the average across five seeds as the final accuracy. Our results are presented in Appendix Figure \ref{fig:evp_ablations}. We noticed that even in cases where the model dimension is sufficiently large to store all necessary information, our modifications lead to $100\%$ accuracy for every hyperparameter set and every seed we use, in contrast with the Based model, where we observe degradation for certain learning rates.

\section{Pile Dataset Experiment Details}
\label{ap:lm}

\begin{table}[!hbtp]
    \centering
    \begin{tabular}{r|c} \toprule 
        Model & \# Parameters \\ \midrule
        Attention & 151 880 448 \\ \midrule
        Based & 147 542 016 \\
        ReBased & 147 548 928 \\ \bottomrule
    \end{tabular}
    \caption{Parameters count for the pile experiment. See Appendix \ref{ap:lm}.}
    \label{tab:lm_parameters}
\end{table}

We train our model on the tokenized Pile dataset published on huggingface hub\footnote{\href{https://huggingface.co/datasets/EleutherAI/pythia_deduped_pile_idxmaps}{https://huggingface.co/datasets/EleutherAI}}. Note that this tokenization differs from the one used in Based\footnote{See \href{https://wandb.ai/hazy-research/zg-neox/reports/Zoology---Vmlldzo2MjExMTc2?accessToken=z1e7bu20nm8r5l8eqp2e9ftork7o94tthd2nq6wimai1k4ger1f3vhb76rfvliw2}{report}}. We also use our pipeline, which we plan to release to the public soon. We do not use rotary positional embeddings \cite{su2024roformer} or other tricks, as we copy our models from the Based repository. Hyperparameters can be found in Table \ref{tab:hp}.

\begin{table}[!h]
    \centering
    \begin{tabular}{|c|c|} \toprule
    Hyper-Parameter & Value\\ \midrule
        warmup steps &  200\\
        max grad norm &  1\\
        num steps &  20000\\
        seq len &  4096\\
        lr &  1e-3\\
        weight decay & 0.1\\
        num heads & 12 \\
        d model & 768\\
        effective batch size & 1024 \\\bottomrule
    \end{tabular}
    \caption{Hyperparameters used for training.}
    \label{tab:hp}
\end{table}

As in \citet{mqar}, we use more hyperparameters in the Based/ReBased models then in the Attention baseline. The number of layers and head dim reported in Tables \ref{tab:hp_attn} and \ref{tab:hp_based}. We use a hybrid architecture for the Based/ReBased models where we use short convolution as a mixer for every odd-numbered layer.

\begin{table}[!h]
    \centering
    \begin{tabular}{|c|c|} \toprule
    Hyperparameters & Value\\ \midrule
        layers & 12 \\
        head dim & 64 \\\bottomrule
    \end{tabular}
    \caption{Attention hyperparameters.}
    \label{tab:hp_attn}
\end{table}

\begin{table}[!h]
    \centering
    \begin{tabular}{|c|c|} \toprule
    Hyper-Parameter & Value\\ \midrule
        layers & 18 \\
        head dim & 16 \\\bottomrule
    \end{tabular}
    \caption{Based/ReBased hyperparameters.}
    \label{tab:hp_based}
\end{table}

\begin{figure*}[!htbp]
    \centering
    \includegraphics[width=\linewidth]{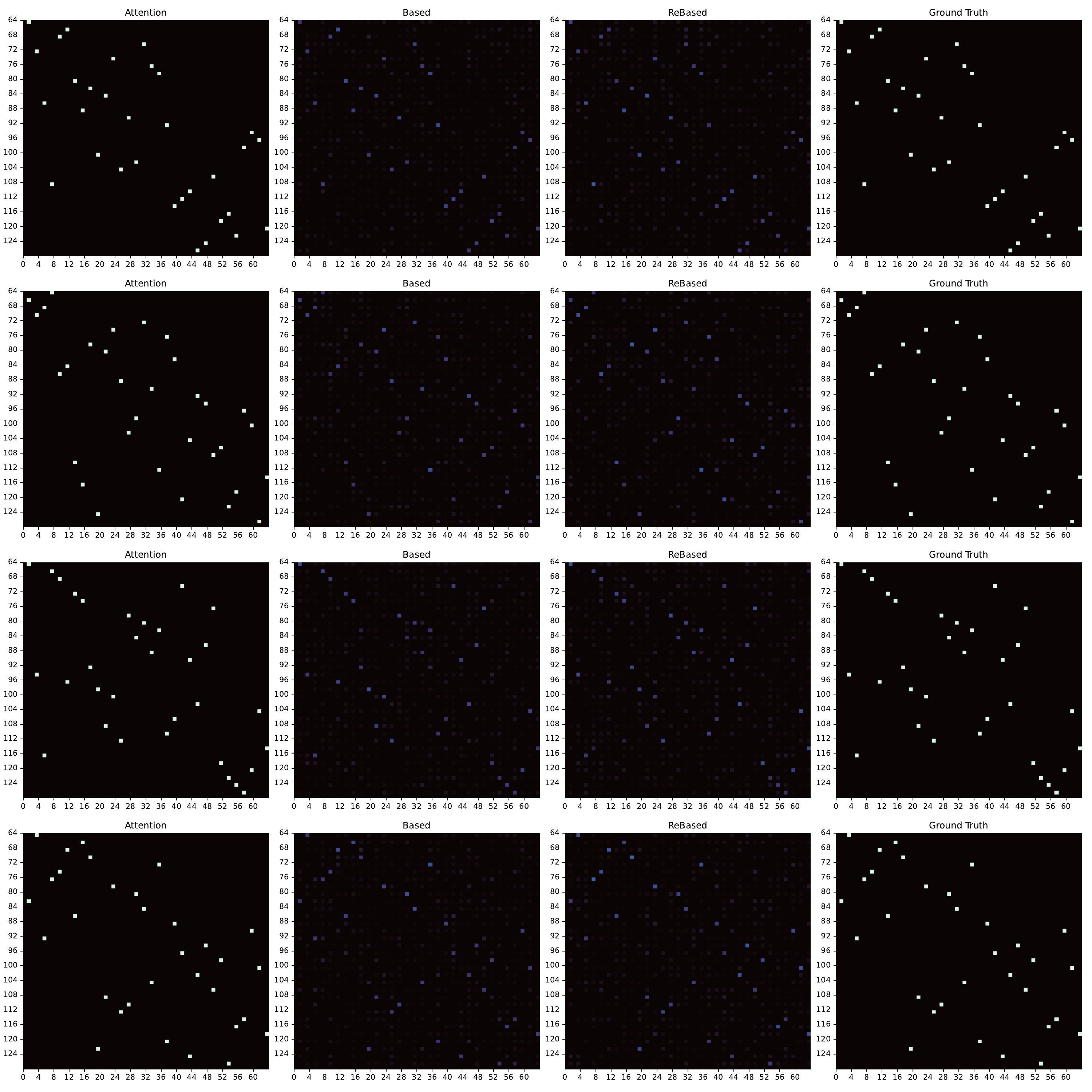}
    \caption{Attention matrix for the different models, and ground truth positions for the query. We measure IoU between the model's attention and ground truth matrix for $10000$ examples. llustration of the experiment is described in Section \ref{sec:analysis} Results are presented in Table \ref{tab:iou}.}
    \label{fig:ap_iou}
\end{figure*}

\section{Additional Analysis}
\label{sec:additional_analysis}

\begin{figure*}

\includegraphics[width=\linewidth]{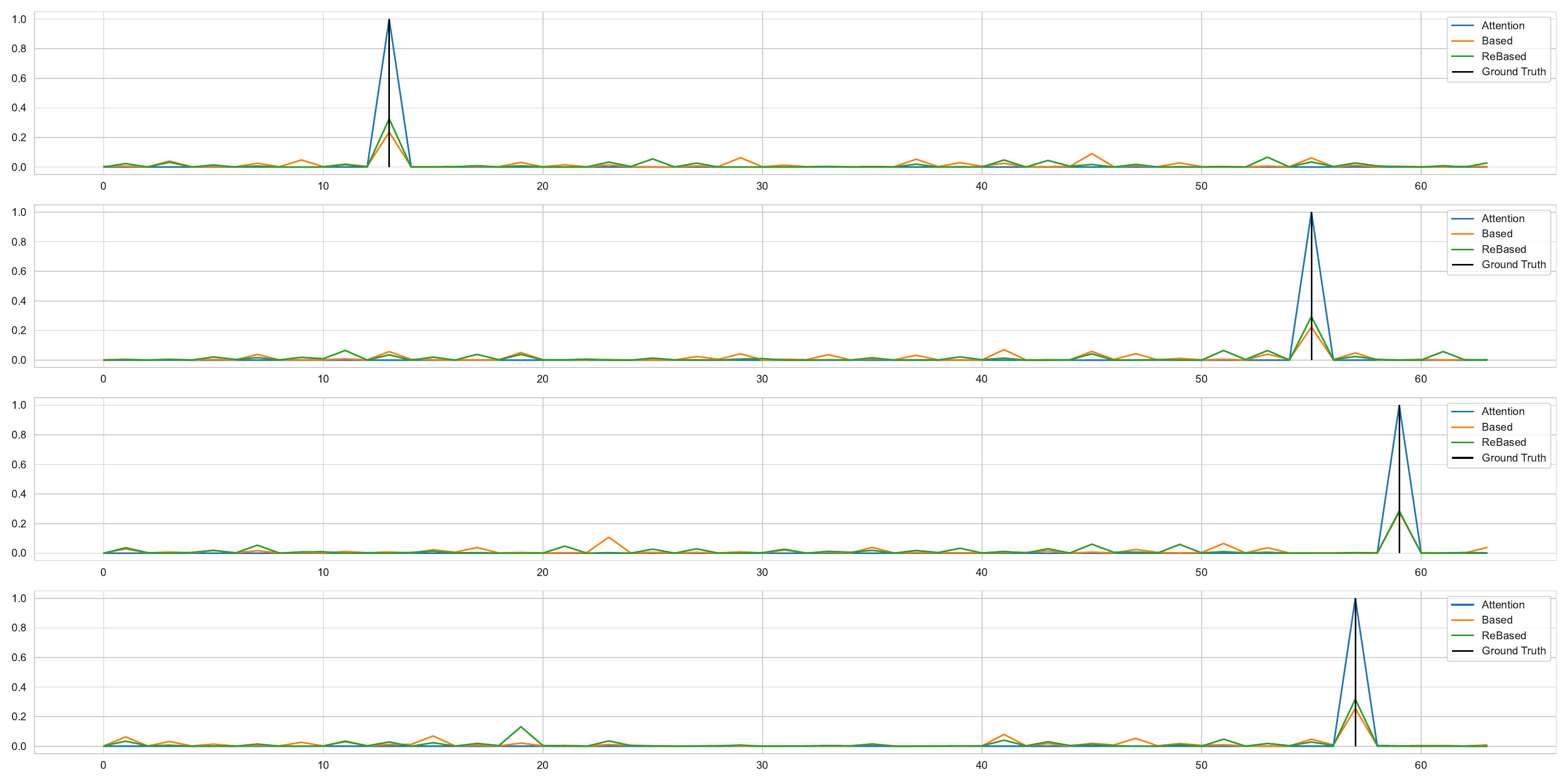}
    \caption{Attention scores for a random example. Based and Rebased scores are noisy, while attention has one peak at the ground truth position. See Appendix \ref{ap:mqar_details}.}
    \label{fig:ap_peaks}
    
\end{figure*}

In this section, we provide additional results and experiments to further understand how the ReBased model learns dependencies. First, we offer more examples for our experiments with attention matrices, as detailed in Section \ref{sec:analysis}. Attention matrices for random examples from the test set are presented in Figure \ref{fig:ap_iou}. Generally, we can observe that attention "fires" more intensely at retrieving tokens compared to Based/ReBased. This result suggests that there may still be a flaw in our kernel function that distributes attention to irrelevant tokens. We further investigate this phenomenon by analyzing the distribution of attention on the last token, as shown in Figure \ref{fig:ap_peaks}. The number of noisy tokens for the ReBased model is smaller compared to Based, but Attention exhibits superior results.

\begin{figure*}[!htbp]
    \centering
    \includegraphics[width=\linewidth]{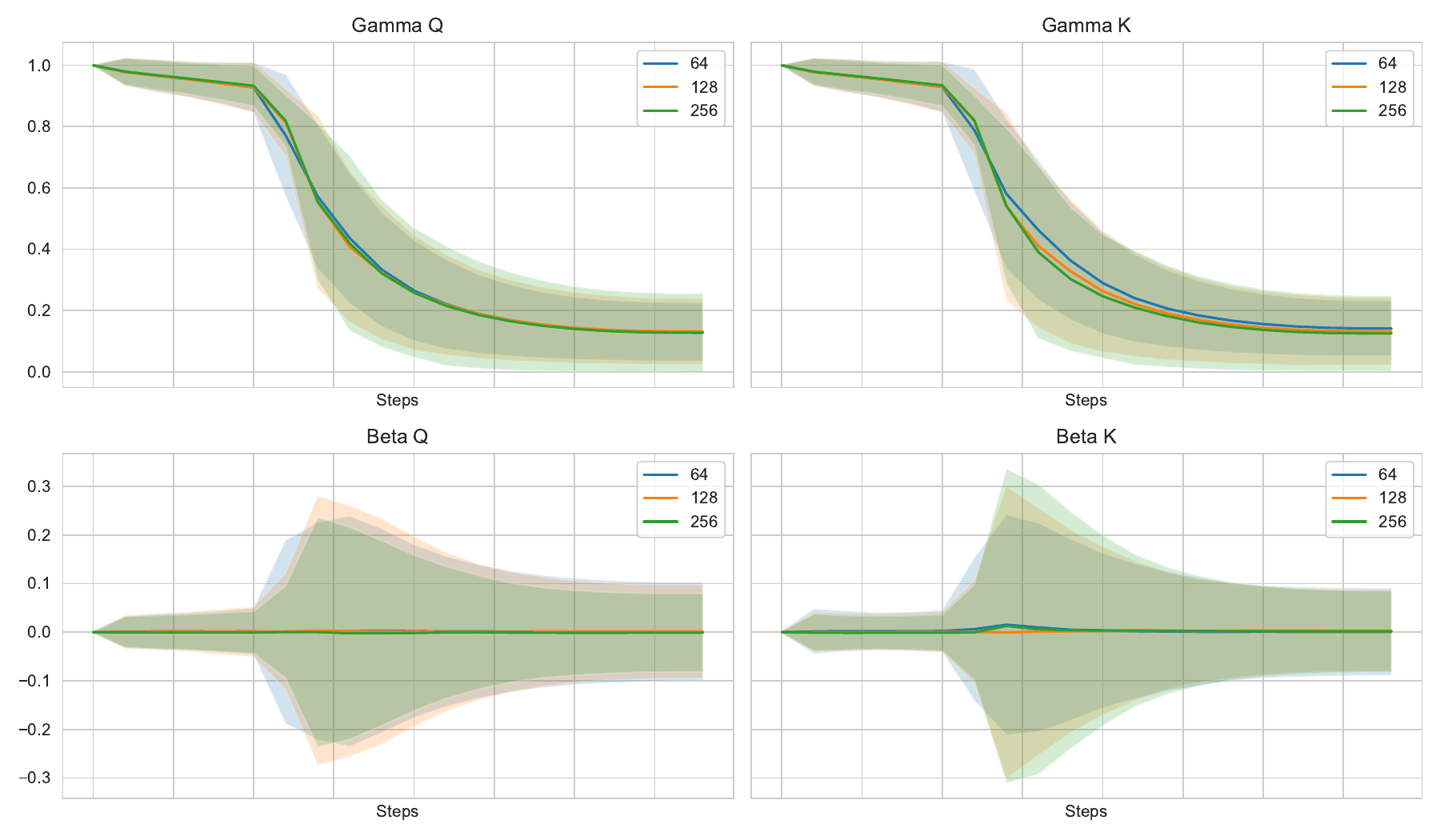}
    \caption{Analysis of the layer normalization parameters. Mean value of the scale parameter (gamma) tends to the same value of about $0.13$, and the bias parameter (beta) tends to $0$. See Section \ref{sec:additional_analysis}.}
    \label{fig:gamma_beta}
\end{figure*}

Layer normalization is the main difference between the Based and ReBased models. Therefore, it is important to analyze the parameters obtained during the training process. We logged the mean and standard deviation of the parameters across different sequence lengths. Our results can be found in Figure \ref{fig:gamma_beta}. Notably, the final parameter value is independent of the training sequence length, which can indicate that we may not need to train the model for all possible lengths. Both $\gamma$ and $\beta$ parameters have high standard deviation values compared to the mean absolute value. Consequently, we can assume that it is important to provide features with different scales.

\end{document}